\documentclass[10pt,conference,a4paper]{IEEEtran}
\usepackage{booktabs} 
\usepackage{float}
\usepackage{color}
\usepackage{graphicx,url,bm}
\usepackage{amsmath}
\usepackage{amssymb}
\usepackage{tabularx} 

\renewcommand{\vec}[1]{\boldsymbol{#1}}
\newcommand{\mat}[1]{\boldsymbol{#1}}

\def\onedot{. }
 
\def\ie{\emph{i.e}\onedot}

\def\etal{\emph{et al}\onedot}
\ifCLASSINFOpdf
\else
\fi

\begin{document}
%
\title{Deep Structured-Output Regression Learning for Computational Color Constancy}

\author{
\IEEEauthorblockN{Yanlin Qian, Ke Chen, Joni-Kristian K\"am\"ar\"ainen}
\IEEEauthorblockA{Department of Signal Processing\\
Tampere University of Technology\\
\url{http://vision.cs.tut.fi/}}
\and
\IEEEauthorblockN{Jarno Nikkanen
}
\IEEEauthorblockA{Intel Finland}
\and
\IEEEauthorblockN{Jiri Matas}
\IEEEauthorblockA{Center for Machine Perception\\
Czech Technical University}
\url{http://cmp.felk.cvut.cz/}
}


%


\maketitle
\begin{abstract}
The color constancy problem is addressed by structured-output regression on the values of the fully-connected layers of a convolutional neural network. The AlexNet and the VGG are considered and VGG slightly outperformed AlexNet. Best results were obtained with the first fully-connected "{\it fc}$_6$" layer and with multi-output support vector regression.
Experiments on the SFU Color Checker and Indoor Dataset benchmarks demonstrate that our
method achieves competitive performance, outperforming the state of the art
on the SFU indoor benchmark. 
\end{abstract}


\section{Introduction}

A visual system possesses color constancy if it perceives colors almost independently of the prevailing illumination for a wide range of conditions. In digital cameras, the so-called automatic white balance  aims at reaching this property.
Color constancy is a desirable property in many computer vision and graphics applications where the intrinsic color of the object is needed -- for accurate classification, regression, segmentation, and  feature extraction~\cite{gijsenij2010generalized} as well as for accurate scene
rendering.  

Methods aiming at achieving color constancy fall into two groups. The first
estimate illumination properties  of the observed scene which is followed by image color correction called chromatic adaption \cite{barnard2000improvements,barnard2002comparison}
\cite{buchsbaum1980spatial},
\cite{chakrabarti2012color},
\cite{finlayson2001color,gijsenij2011color,van2007edge}.
The second group, operates on an illumination invariant representations without explicitly estimating the scene illumination. The invariant features are designed or learned to depend only on reflectance characteristics or spatial structure
\cite{arandjelovic2012colour,finlayson2001hue,lee2014taxonomy,van2005edge}. 

\begin{figure}
\begin{center}
\includegraphics[width=\linewidth]{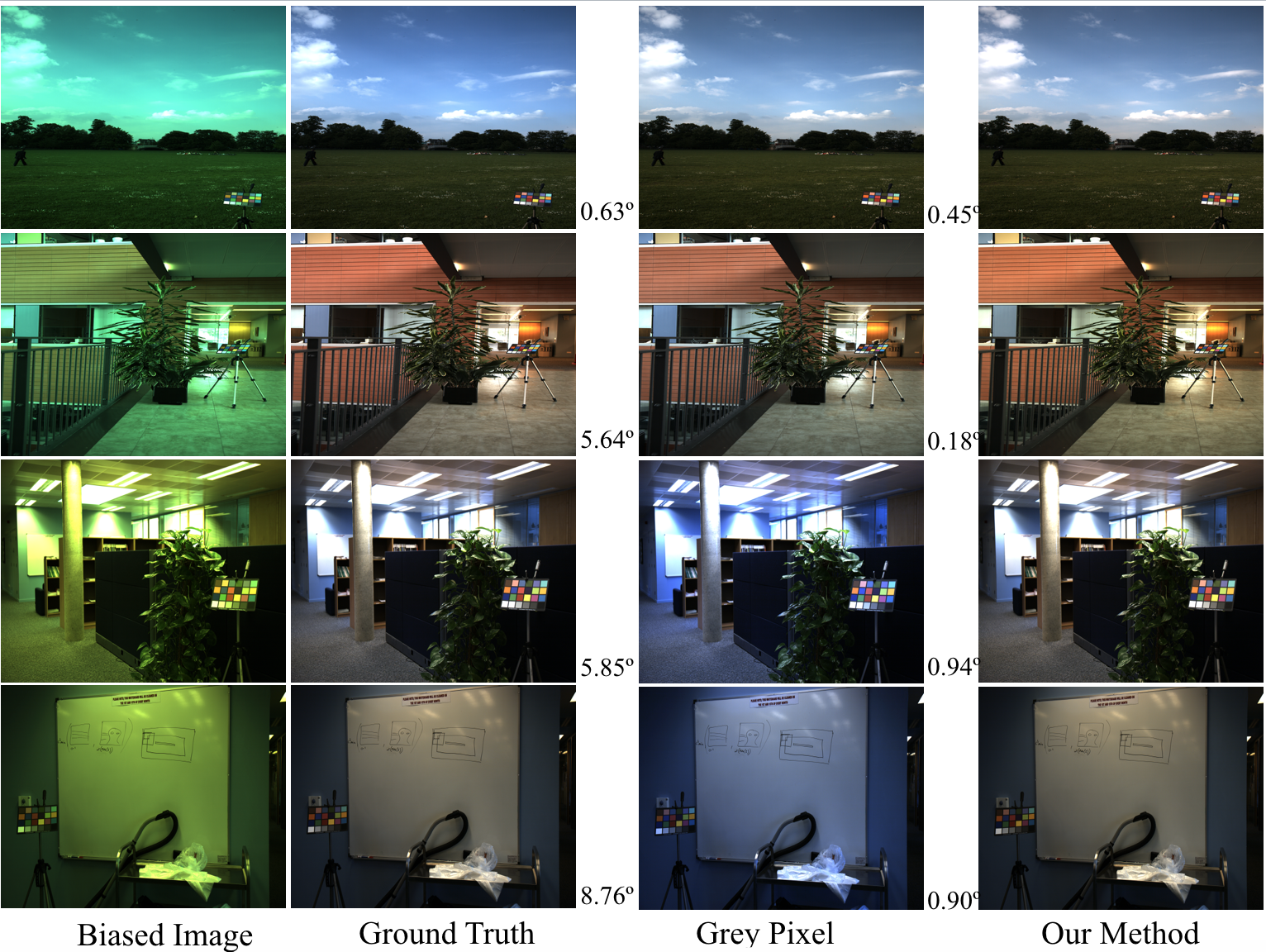}
\caption{Image Correction of examples of the SFU Color Checker Dataset using different approaches.\label{fig:intro}}
\end{center}
\end{figure} 

We propose a novel method for estimating scene illumination color that uses
a structured-output regression on the output of a single fully-connected layer from a deep net.  Image color is then corrected on the basis of  the estimated illumination parameters. The deep net extract visual information that has been shown to facilitate color constancy \cite{bianco2015color} as well as other computer vision task  
\cite{kavukcuoglu2010learning,simonyan2014very,Szegedy_2015_CVPR}.

The structured-output regression models the cross-channel correlations, which is the main contribution of our work.
In the literature \cite{funt2004estimating}, single-output regressor, \ie support vector regression, has been employed to learn the relationship between
observation variables and each color dimension of the target variables
independently. 

We show experimentally that the proposed framework achieves competitive
performance comparable with several state-of-the-art methods on two popular
color constancy benchmarks.  

\begin{figure*}[thbb]
\begin{center}
\includegraphics[width=0.88\linewidth]{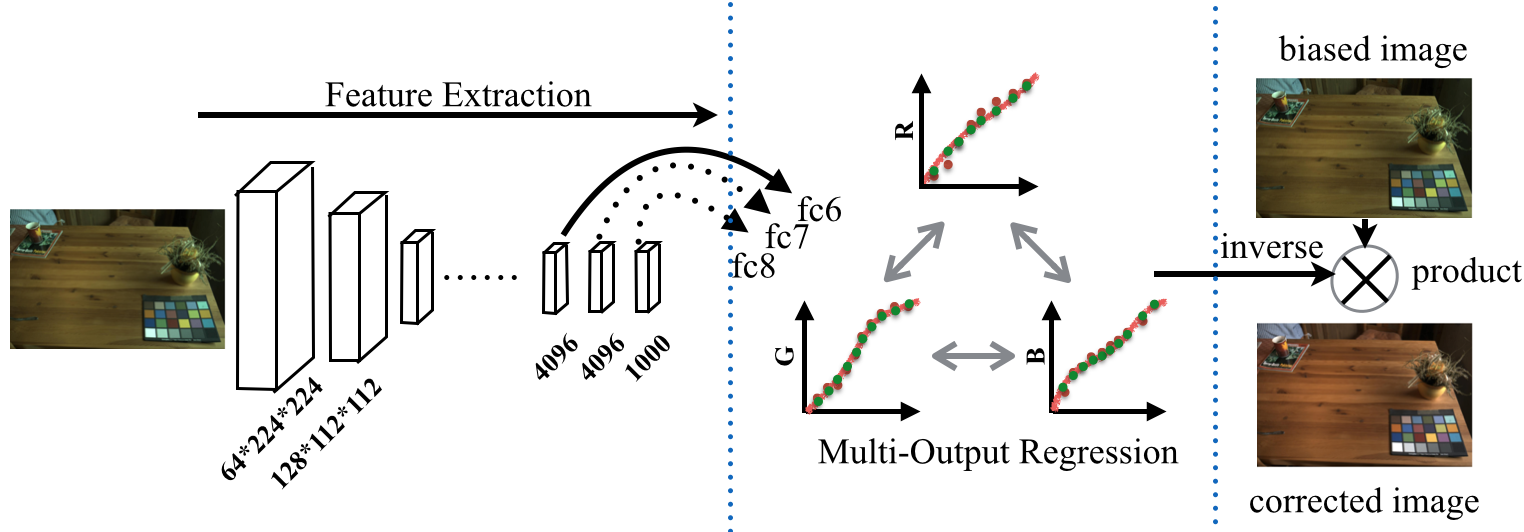}
\caption{The proposed color constancy method consists of CNN feature extraction, multi-output regression learning and color correction.\label{fig:framework}}
\end{center}\vspace{-0.25cm}
\end{figure*}

\section{Related Work}

Color constancy is a well-established problem. To understand the complexity of the
problem, we revisit one of the simplest but yet often adopted model of  image formation, i.e,  the of  Lambertian surfaces with purely diffuse reflection lit by constant scene  illumination. In this case, the measured image color values $\rho_{i}(x,y)$ at pixel $x,y$ is a function of  the global light source $I(\lambda)$, the spectral response of the  sensors  $S_{i}(\lambda)$, and the surface
reflectance $R(x,y,\lambda)$, $\lambda$ denotes the wavelength. The dependence is captured by the following formula: 
\begin{equation*}
\rho_{i}(x,y) = \int I(\lambda)S_{i}(\lambda)R(x,y,\lambda) d\lambda, i\in 
  \{\mbox{R,G,B}\}.
\label{eq:formation}
\end{equation*}
The formulate expresses that fact that the RGB value of each pixel is  obtained by integration over wavelengths with different weights in each single channel. 

In a color constancy task, $\rho$ is given, while $\rho _{gt}$ under canonical illumination is sought. We consider the case when neither camera sensibility $S(\lambda)$ nor the illumination $I$ are available, making this task under-constrained or under-determined \cite{funt1998machine}.


 Below, a basic introduction is given to illumination estimation algorithms.
To make different assumptions unified, Van de
Weijer \etal \cite{van2007edge} proposed a formulation, which can cover
different algorithms based on exploiting imagery statistics in a
single image to estimate the scene illumination $I$ as the following: 
\begin{equation}
I(n,\rho,\sigma) = K\sqrt[\rho]{\int\int {(\rho(x,y)\otimes G_\sigma(x,y))}^{\rho}dx dy}, 
\label{eq:formation}
\end{equation}
where illumination $I$ is decided by the order $n$ of the derivative,
Minkowski-norm $\rho$, and the scale parameter $\sigma$ of a Gaussian
filter. Operator $\otimes$ defines convolution between the image
$\rho$ and Gaussian filter $G$. $K$ is a constant added to make $I$ in
unit length. By varying
($n,\rho,\sigma$) in Eq. (\ref{eq:formation}), a number of existing
algorithms are generated under different assumptions. For instance,
GM-pixel \cite{forsyth1990novel}, GW \cite{buchsbaum1980spatial}, SoG
\cite{finlayson1994color}, gGW \cite{van2007edge} and WP
\cite{cardei1999white} are clustered as one group -- zero-order
statistics based methods, as they are all under the assumption that
a part of images (\textit{e.g.}, local regions) has gray average
color.  
In terms of higher order statistics, first-order \cite{van2007edge}
and second-order based methods \cite{van2007edge} respectively adopt
the assumption that edges or gradient of edges has gray average color
intrinsically. In the light of this, illumination is measured by the offset
of the average color. Gamut-mapping based methods assume that in real
world, part of color spectral distribution of objects is able to be observed
\cite{barnard2000improvements,forsyth1990novel}. In other
words, the limited set of colors in each biased image is caused by
a certain lighting condition, which encourages researchers to
introduce a learning based framework to recognize canonical illumination pattern with a sufficient amount of labeled training data.  


Machine learning algorithms have succeeded in learning a function directly mapping  low-level image feature to illumination. Support Vector Regression (SVR) is first applied to deal with color constancy task in
\cite{funt2004estimating}, but the performance is limited by weak
predictive power of the color histogram. Bayesian approaches
\cite{gehler2008bayesian} learn the probability of illumination assuming  normal-distributed reflectances. The Exemplar method \cite{joze2014exemplar} estimates illumination via finding nearest neighbor surfaces of a test image using  an unsupervised clustering of texture and color features, 

CNNs have been applied successfully to the color constancy problem. Barron \etal \cite{barron2015convolutional} formulate the color constancy as a 2D spatial localization task in log chromatic space using a convolutional classifier specifically designed for object localization. In Bianco \etal \cite{bianco2015color}, a five-layer ad-hoc CNN is designed that combines feature generating and multi-channel regression to estimate illumination in an end-to-end way, similarly to \cite{van2007using}. Bianco \etal \cite{bianco2015color} compare results obtained with their five-layer ad-hoc CNN with reference -- the output of  layer {\it fc}$_6$ of an AlexNet fed into SVR. The
AlexNet with SVR performed worse, but outperform most of the statistic-based methods. The 
selection of AlexNet and layer {\it fc}$_6$ is ad-hoc, no other layers or net architectures were explored.

\section{Methodology}

The proposed deep structured-output regression  color constancy method consists of three steps (see Fig. \ref{fig:framework}):
\begin{enumerate}
\item Calculation of the response, after the non-linearity,  of  the fully-connected {\it fc}$_6$ layer of the VGG CNN \cite{simonyan2014very}.
\item Illumination estimation by structured-output regression.
\item Image correction with the illumination obtained.
\end{enumerate}
Below we describe the three steps in details.

\subsection{CNN Feature Extraction}\label{Sec:feat}

Color constancy regression has been shown to perform well \cite{bianco2015color,funt2004estimating}. CNN is a robust regressor for a number of problems \cite{gijsenij2011computational}. 

The nets showing consistent performance for a number of vision problems are VGG \cite{simonyan2014very,Szegedy_2015_CVPR} and AlexNet \cite{krizhevsky2012imagenet}. We tested both architectures and the VGG outperformed AlexNet in all experiments (Table \ref{tab:vggvsalexnet}). The 19-layer VGG CNN we adopted has the same  structure as in \cite{simonyan2014very}.

We did not attempt training the VGG from scratch since the size of the available training set was limited to about 1000 samples (568 from the SFU Color Checker Dataset \cite{gehler2008bayesian,shi2010re} and 321 from the SFU Indoor Dataset \cite{barnard2002data}).
We therefore tried only to fine-tune the VGG CNN that was pre-trained on ImageNet.  However, the fine-tuned net performed worse than the original pre-trained VGG.
Therefore, the original VGG CNN without fine-tuning was used
in all experiments.

We experimented with features from different fully-connected layers, namely ``{\it fc}$_6$'', ``{\it fc}$_7$'' (both having 4096 dimensions) and ``{\it fc}$_8$'' (1000 dimensions). The results are presented in Table \ref{tab:cnnlayers}.  CNN features from layer ``{\it fc}$_6$'' were selected as the performed the best, and can be extracted the fastest.
The CNN feature is extracted from an image resized to $224\times
224$.


\subsection{Structured-Output Regression Learning}\label{Sec:mr}

The regressor is trained on the  set $\{(\vec{x,y})\}_i$, $i=1,2, \cdots, N$ includes vector-valued $\vec{y}$,
where $\vec{x}$ denotes the 4096-dimensional VGG ``{\it fc}$_6$'' feature vector.
The objective of structured-output
\cite{funt2004estimating} is to learn regression functions
independently for ${y}^l, l=1,2,3$ where $i$ denotes the color (RGB)
dimensions. The single-output regression learning is formulated as:  
\begin{equation}
\min ~~~\dfrac{1}{2}||\vec{w}||_2^{2}+C\sum_{i=1}^N \text{loss}({y}^l_i, f(\vec{x}_i)),\label{eqn.sobj}
\end{equation} 
where $\vec{w} \in \mathbb{R}^d$ is the weight vector to be optimized,
the parameter $C$ controls the regularization trade-off and
$f(\vec{x}_i)=\phi(\vec{x}_i)^T \vec{w} + 
b$ with $\phi(\cdot)$ is the kernel function to project $\vec{x}$ to a
high-dimensional Hilbert space.   

In order to take the possible correlations between output
variables into account, general structured-output regression has the following formulation: 
\begin{equation}
\min ~~~\dfrac{1}{2}\sum_{l=1}^3||\vec{w}^l||_2^{2}+C\sum_{i=1}^N \text{loss}(\vec{y}_i, F(\vec{x}_i)),\label{eqn.mobj}
\end{equation}
where  $F(\vec{x}_i)=\phi(\vec{x}_i)^T \mat{W} +
\vec{b}$ with $\mat{W} \in \mathbb{R}^{d\times 3}$ and $\vec{b}\in
\mathbb{R}^3$. Eq. (\ref{eqn.mobj}) is the general formulation for
multi-output regression. Adopting different loss functions leads to different regressors. 
Given training samples $\{(\vec{x}, \vec{y})\}_i,
  i=1,2,\cdots,N$, a multi-output regressor (\textit{i.e.},
  multi-output support vector regression (MSVR) and multi-output ridge
  regression (MRR)) is employed to learn both the input-output
  relationship and latent correlations across output
  variables jointly.  

\vspace{\medskipamount}\noindent\textbf{Multi-output ridge regression   (MRR) -- }
minimizes
quadratic loss function:
\begin{equation}
\min ~~~\dfrac{1}{2}\sum_{l=1}^3||\vec{w}^l||_2^{2}+C\sum_{i=1}^N \|\vec{y}_i - (\phi(\vec{x}_i)^T\mat{W} + \vec{b} )\|_2^2.\label{eqn.rr}
\end{equation}
It has a closed-form solution based on matrix
inversion \cite{borchani2015survey,chen2013cumulative}.  

\vspace{\medskipamount}\noindent\textbf{Multi-output support vector regression (MSVR) -- }
with a
$\epsilon$-sensitive loss function for support vector regression (SVR)
\cite{smola2004tutorial},  
\begin{equation*}
\text{loss}({y}^l_i, f(\vec{x}_i)) = 
\begin{cases}
0,~~~~~~~~~~~~~~~~~~~\text{if}~~|{y}^l_i - f(\vec{x}_i)| < \epsilon\\
|{y}^l_i - f(\vec{x}_i)| - \epsilon, ~~\text{if}~~|{y}^l_i - f(\vec{x}_i)| \geqslant \epsilon
\end{cases}
\end{equation*}
is solved by using cutting-plane strategies \cite{joachims2009cutting}. $\epsilon$ in formulation controls the insensitivity to output bias.

%

\begin{table*}[!t]
\centering
\caption{Median, Mean and Max angular errors for state-of-the-art algorithms on two benchmarking datasets.}
\label{tab:stateoftheart}
\begin{tabular}{lrrrrrrrr}
\toprule 
 & & & \multicolumn{3}{c}{SFU Color Checker} & \multicolumn{3}{c}{SFU Indoor} \\
  & statistic-based & learning-based & Median & Mean & Max  & Median & Mean & Max\\
 \midrule
Doing Nothing (DN) & -- & -- &13.55 & 13.63 & 27.37 &15.60 & 17.30& -- \\ 
 White Patch (WP) \cite{brainard1986analysis}& ${\surd}$ &-- & 5.61& 6.27& 40.59 & 6.50 & 9.10 & 36.20 \\ 
 Gray World (GW) \cite{buchsbaum1980spatial} & ${\surd}$ & -- &6.27 & 6.27& 24.84 & 7.00 & 9.80 &37.30 \\ 
 Shades of Gray (SoG) \cite{finlayson2004shades}& ${\surd}$ &-- &4.04 &4.85 & 19.93 & 3.70& 6.40 & 29.60\\ 
 general Gray World (gGW) \cite{barnard2002comparison} & ${\surd}$ & --&3.45 & 4.60 & 22.21 & 3.30 & 5.40 &28.90 \\ 
first-order Gray Edge ($1^{st}$GE) \cite{van2007edge}  & ${\surd}$ & --&4.55 &5.21 &19.69 &3.20 &5.60 &31.60 \\ 
second-order Gray Edge ($2^{st}$GE) \cite{van2007edge} & ${\surd}$ & --&4.43&5.01 &16.87 &2.70 &5.20 &26.70 \\ 
Bright Pixels \cite{joze2012role}& ${\surd}$ & --&--&-- &-- &2.61 &3.98 &--\\ 
 Grey Pixel (std) \cite{yang2015efficient}& ${\surd}$ & --&3.20&4.70 &-- &2.50 &5.70 &--\\ 
 Grey Pixel (edge) \cite{yang2015efficient}& ${\surd}$ & --&3.10&4.60 &-- &2.30 &5.30 &--\\ 
 Weighted Gray Edge (WGE) \cite{gijsenij2012improving}& ${\surd}$ &${\surd}$ &-- &-- &-- & 2.40& 5.60& 43.80 \\  
\midrule
Gamut Mapping (GM-pixel) \cite{barnard2000improvements}& ${\surd}$ & ${\surd}$ & { 2.30} & 4.20 &23.20 &2.30  & 3.70  &27.10 \\ 
Gamut Mapping (GM-edge) \cite{barnard2000improvements} & ${\surd}$ & ${\surd}$ & 5.00 & 6.50 &29.00 &2.30 &3.90 & 29.70 \\ 
\midrule
Bayesian (BAY) \cite{gehler2008bayesian} &-- &${\surd}$ &3.44 &4.70 &24.47 &-- &-- &--  \\ 
Natural Image Statistics (NIS) \cite{gijsenij2011color}&-- &${\surd}$ &3.13 &4.09 &26.20 &-- &-- &--  \\ 
Spatial-spectral (SS-GenPrior) \cite{chakrabarti2012color}&-- &${\surd}$  &2.90 &3.47 &{\bf 14.80}&-- &-- &--  \\ 
Spatial-spectral (SS-ML) \cite{chakrabarti2012color}&-- &${\surd}$& 2.93&3.55 &15.25 & 3.50 & 5.60 & -- \\ 
Exemplar \cite{joze2014exemplar} & -- &${\surd}$ &{\bf 2.30} &{\bf 2.90} &19.40 &-- &-- &--  \\ 
SVR \cite{funt2004estimating} &-- &${\surd}$ &6.67 &7.99 &26.08 & 2.20 & --&--  \\ 
\textbf{CNN+RR} & -- &${\surd}$ & 4.08 & 5.30  & 22.69 & 1.99 & 3.32 & 19.46\\ 
CNN+SVR \cite{bianco2015color}  & -- &${\surd}$ & 3.09 &4.74 &29.15 &-- &-- &--   \\ 
\textbf{CNN+MRR}& -- &${\surd}$ & 3.76 & 4.87 &21.68 &1.93 &{\bf 3.24} &{\bf 18.89} \\
\textbf{CNN+MSVR}& -- &${\surd}$ & 2.68 & 4.29 & 20.35 &{\bf 1.64} &{ 3.25} &21.39 \\
\bottomrule
\end{tabular}\vspace{-0.25cm}
\end{table*}

\subsection{Image Correction \label{subsec:imagecorrection}}

Given an unseen test image, an estimate of global illumination is
obtained with the trained structured-output regressor in Section
\ref{Sec:mr}. By one more step, we can recover unbiased image with it,
which is called chromatic adaptation \cite{von1902}. Among many
chromatic adaptation methods (\textit{e.g.} Bradford
\cite{hunt2011metamerism} and CIECAT02 \cite{moroney2002ciecam02}), we
choose to use the von Kries model \cite{von1902}: $I=W\times L$ (thus
$L=W^{(-1)}\times I$), which is based on the simplified  assumption
that each channel of color is modified separately to model
photo-metric change and the sensors $S(\lambda)$ in Eq. (\ref{eq:formation}) are narrow-band approaching a delta function. Despite its simplicity the model is relatively stable in practice
\cite{brill2008minimal}.

\section{Experiment \label{section:result}}

\begin{figure}
\begin{center}
\includegraphics[width=\linewidth]{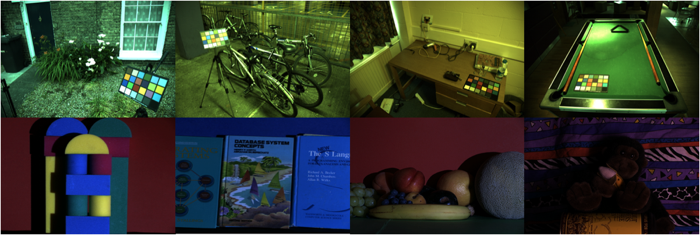}
\caption{Images from the two benchmarks used in the experiments. Top: the SFU Color Checker Dataset; Bottom: the SFU Indoor Dataset.\label{fig:datasets}}
\end{center}\vspace{-0.75cm}
\end{figure} 

\subsection{Datasets and Measurement}

The proposed method is evaluated on two popular benchmarking
datasets \cite{barnard2002data,gehler2008bayesian,shi2010re} with linear images. The SFU Color Checker dataset \cite{gehler2008bayesian,shi2010re} contains 568 14-bits dynamic range images which all include the Macbeth Color Checker chart. The SFU Indoor dataset \cite{barnard2002data} includes more artificially-looking scenes, containing 321 images captured in 11 different controlled lighting conditions. Groudtruth of both datasets are provided. Sample images from the two datasets are shown in Fig. \ref{fig:datasets}.

Following the standard procedure \cite{bianco2015color}, we evenly split the data into training, validation and testing sets. We train our regressors and use the validation set to tune free parameters of the regressors. The procedure is repeated $30$ times with different random splits. The average errors are reported. 

Following the prior work \cite{barnard2002data,ciurea2003large,gehler2008bayesian,shi2010re}, 
the accuracy of illumination estimation is measured by the angular error $\varepsilon$ between estimated illumination $I$ and groundtruth~$I_{gt}$:
\begin{equation*}
\varepsilon_{I,I_{gt}} = \arccos(\frac{I\cdot I_{gt}}{\parallel I\parallel \cdot \parallel I_{gt}\parallel}), \enspace 
\label{eq:metric}
\end{equation*}
where $\cdot$ denotes the inner product between vectors, $\parallel \cdot\parallel$ is Euclidean norm.%
\footnote{We are aware that error measure is not directly related to the appearance of an image corrected by the estimated illumination $I$. We follow the standard procedure --  discussion of color error metrics is the out of scope of the paper.
}


\subsection{Regressor Learning Settings}
Free parameters in the aforementioned regressors 
need to be tuned: regularization trade-off parameter $C$, kernel coefficient $\gamma$ for the RBF kernel and insensitivity parameter $\epsilon$ in MSVR. 
\begin{itemize}
\item For multi-output ridge regression, only trade-off parameter $C$ is chosen from $C \in 10^{(-2:1:2)}$.\footnote{Following the usage in Matlab, the notation (x:y:z) represents an array starts from x to z with the step of y.}
\item For our structured-output SVR implementation (i.e., CNN+MSVR), we choose the optimized parameters by searching in grid space where $C\in 10^{(-3:1:5)}$, $\gamma \in 10^{(-4:1:4)}$ and $\epsilon \in 10^{(-4:2:3)}$.
\end{itemize}

\subsection{Comparison to State-of-the-Art \label{subsec:result}}

In Table \ref{tab:stateoftheart}, the median, the average, and the maximum of the angular errors of state-of-the-art methods and our methods are evaluated and compared. 
We categorize these methods into three groups in the table: non-learning algorithms (top), camera-specific ones (middle) and learning-based ones (bottom). The best results for each metric are in bold. 

The multi-output ridge regression (CNN+MRR), multi-output support vector regression (CNN+MSVR) and  ridge regression (CNN+RR) are our implementations. They all share the same CNN features from the ``{\it fc}$_6$" layer of VGG CNN. 
From Table \ref{tab:stateoftheart}, it is evident that even a medium-depth CNN \cite{krizhevsky2012imagenet} coupled with SVR (CNN+SVR \cite{bianco2015color}) shows better performance than all statistics based algorithms and most of gamut based and learning based methods. 
Specifically, our CNN+MSVR outperforms the best non-learning based methods (Grey Pixel) on the SFU color checker and the SFU indoor datasets by decreasing the median angular error with 13.55$\%$ and 28.70$\%$ respectively. Compared to the existing camera-specific and learning based frameworks, the significant improvement of our CNN-MSVR is achieved on the SFU Indoor dataset, while our solution is comparable to the state-of-the-art on the SFU Color Checker benchmark. 

The significant improvement on CNN+SVR over its direct competitor SVR can be attributed to the introduction of powerful CNN features, which demonstrates the advantages of adopting CNN features for illumination estimation.
Moreover, direct comparison between CNN+SVR and CNN+MSVR substitutes our main contribution to capture inter-channel correlations by using structured-output regressors, with CNN+MSVR significantly outperforming CNN+SVR. A similar situation is observed for single-output CNN+RR and structural output model CNN+MRR on both benchmarks.


\subsection{Choosing CNN Architecture, Layer, and The Regressor}

\begin{table}[tb]
\centering
\caption{Comparison of deep features from AlexNet and VGG CNNs.}
\label{tab:vggvsalexnet}
\begin{tabularx}{0.5\textwidth}{lrrrrrr} 
\toprule 
 & \multicolumn{3}{c}{SFU Color Checker}  & \multicolumn{3}{c}{ SFU Indoor}\\
 \midrule
 & {Median} & {Mean} &{ Max}  & {Median} & {Mean} &{ Max} \\
 \midrule
AlexNet$_{{\it fc}6}$  & 3.60 & 5.00 & \textbf{19.91} & 1.68 & 3.59 & \textbf{12.96}\\ 
VGG$_{{\it fc}6}$  & \textbf{2.68} & \textbf{4.29} & 20.35 & \textbf{1.64} & \textbf{3.25} & 21.39\\ 
\bottomrule
\end{tabularx}
\end{table}

\begin{table}[t]
\centering
\caption{Comparison of deep features from ``{\it fc}$_6$'', ``{\it fc}$_7$'', ``{\it fc}$_8$'' layers.}
\label{tab:cnnlayers}
\begin{tabular}{lrrr}
\toprule 
 & \multicolumn{3}{c}{SFU Color Checker}\\
 \midrule
 & {Median} & {Mean} &{ Max} \\
 \midrule
VGG$_{{\it fc}6}$  & \textbf{2.68} & {4.29} & \textbf{20.35}\\ 
VGG$_{{\it fc}7}$  & 2.91 & \textbf{4.28} & {23.29} \\ 
VGG$_{{\it fc}8}$  & {3.10} & {4.56} & {26.26} \\
\bottomrule
\end{tabular}
\end{table}

Our best result (last row in Table \ref{tab:stateoftheart}) is obtained by an optimized combination of the CNN architecture, the CNN layer and the regressor, on which evaluation experiments have been done respectively.

Firstly we choose the VGG features as default setting because VGG outperforms AlexNet on two datasets used in our experiments. Table \ref{tab:vggvsalexnet} shows in the metric of median error and mean error, VGG$_{{\it fc}6}$+MSVR achieved 0.7$\sim$0.9 reduction on SFU Color Checker, while on SFU Indoor, this superiority seems not that distinct.

In the second step we investigate the effect of depth of feature extraction layer designed for visual recognition on illumination estimation. The deep feature has been extracted from three fully-connected layers: ``{\it fc}$_6$'', ``{\it fc}$_7$'', ``{\it fc}$_8$'' shown in Fig. \ref{fig:framework}. 
Each type of deep feature is fed into MSVR model to compare the estimation performance. In Table \ref{tab:cnnlayers}, one trend is discovered: the deeper the CNN layer we adopt, the worse the performance we can achieve. 
One possible explanation could be that illumination estimation relies more on low-level information like edges, colors than abstract high-level knowledge for visual understanding as some synthetic examples look fragmented and contain no object or just part of it. Then, the regressor (MSVR) with the highest performance is chosen from three candidates (MSVR, MRR, RR) using the same settings, which is already analyzed in Section \ref{subsec:result}.

\begin{table}[tb]
\centering
\caption{Multi-output SVR training with data augmentation.}
\label{tab:overfitting}
\begin{tabular}{llrrrr}
\toprule 
 & & \multicolumn{4}{c}{SFU Color Checker}\\
 \midrule
 Method & {Augmentation} &{Min} & {Median} & {Mean} &{ Max} \\
 \midrule
  
 CNN+MSVR & -- & -- & \textbf{2.68} &4.29 & \textbf{20.35} \\
 CNN+MSVR & random patches & 0.02 & 3.28 &4.35 &20.42 \\
 CNN+MSVR & regular patches & 0.13 & 3.15 & \textbf{3.87} &21.36 \\
\bottomrule
\end{tabular}
\end{table}

It is known that CNN can work for regression problems directly and we do one more experiments to illustrate the advantages of CNN feature with structured-output regressors over CNN model only.
For the aim to expand the amount of training examples massively, two simple cropping or patching strategies are considered. 
Randomly cropping is to randomly crop 224$\times$224 patches from resized original image ($max(w,h)=1000$), while 
sliding-window patching is aimed to get 224$\times$224 clipped image patches using a square sliding window to scan the whole image region. As shown in Table \ref{tab:overfitting}, this two strategies suffer from over-fitting and cannot achieve competitive performance, compared to the proposed deep structured-output regression frameworks.

\section{Conclusions}

\begin{figure}[t]
\begin{center}
\includegraphics[width=\linewidth]{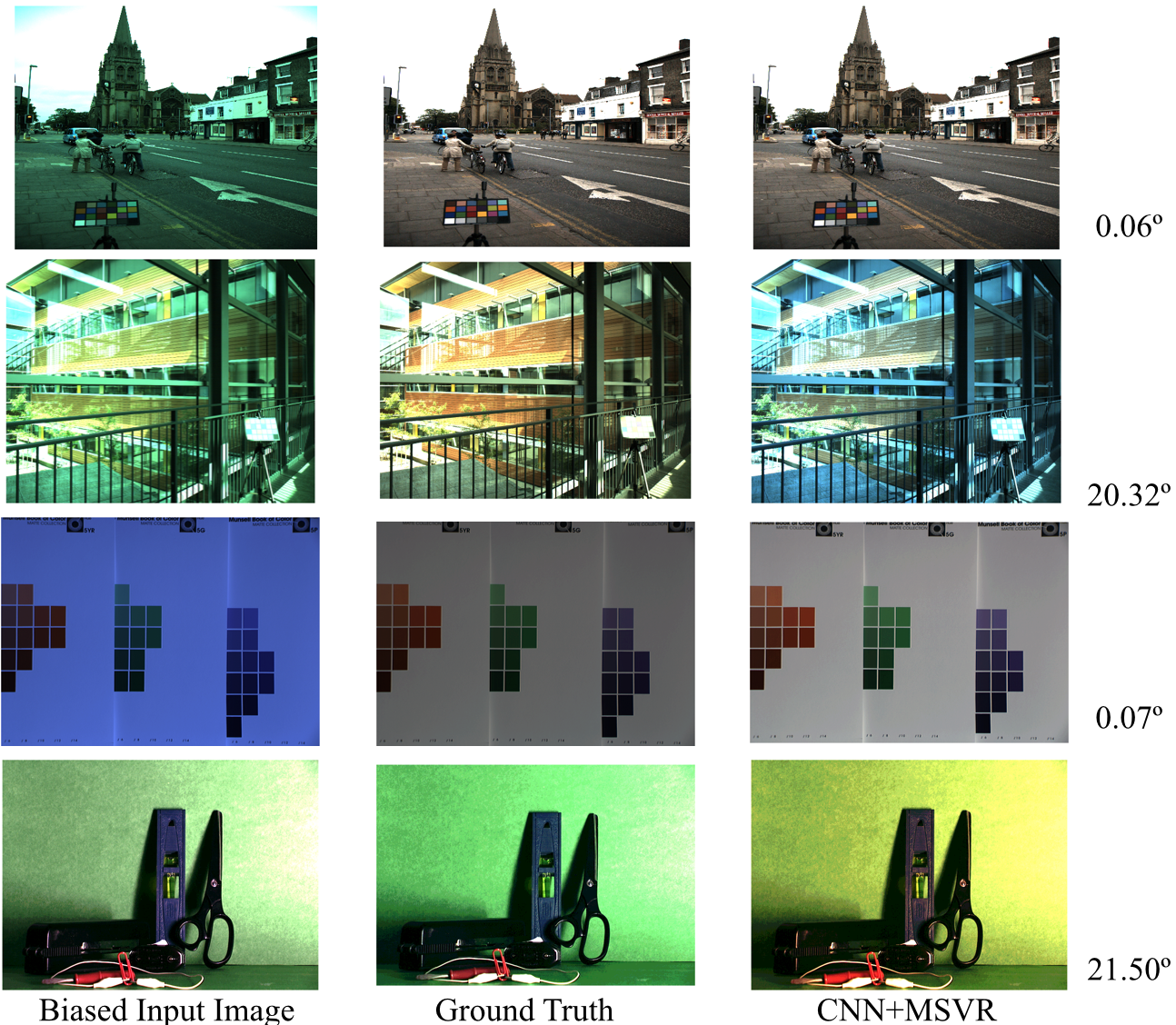}
\caption{Results of the  best and the worst cases on two datasets (first two rows: the SFU Color Checker dataset, last two rows: the SFU Indoor dataset) after applying the proposed algorithm. The angular error between corrected images and groundtruth is marked on the right hand side. \label{fig:example}}
\end{center}
\end{figure} 

A novel method addressing the color constancy problem by structured-output regression on the values of the fully-connected layers of a convolutional neural network has been proposed. Two widely used AlexNet and  VGG CNNs were compared and VGG outperformed AlexNet by a margin of around 30\% in the median error. 
Experiments on the SFU Color Checker and Indoor Dataset benchmarks demonstrate that our
method achieves competitive performance, outperforming the state of the art
on the SFU indoor benchmark with an angular error of $1.64^{\circ}$. 

\section*{Acknowledgment}

This work was funded by the Academy of Finland Grant No.
267581 and 298700, and the Finnish Funding Agency for Innovation (Tekes)
project ``Pocket-Sized Big Visual Data''. J. Matas was supported by the Technology Agency of the Czech Republic project TE01020415 (V3C -- Visual Computing Competence Center).
The authors also acknowledge CSC - IT Center for Science, Finland
for generous computational resources and Intel Finland for technical support.

\bibliographystyle{abbrv}
\bibliography{colorconstancy}

\end{document}